\documentclass{article} % For LaTeX2e
\usepackage{iclr2026_conference,times}

\usepackage{amsmath,amsfonts,bm}

\def\eqref#1{equation~\ref{#1}}
\def\1{\bm{1}}

\DeclareMathAlphabet{\mathsfit}{\encodingdefault}{\sfdefault}{m}{sl}
\SetMathAlphabet{\mathsfit}{bold}{\encodingdefault}{\sfdefault}{bx}{n}

\usepackage{hyperref}
\usepackage{url}
\usepackage{booktabs}
\usepackage{graphicx}
\usepackage{array}
\usepackage[table]{xcolor}
\usepackage{xspace}
\usepackage{amssymb}
\usepackage{wrapfig}
\usepackage[most]{tcolorbox}

\newcommand{\name}[0]{SkillRise\xspace}

\title{SkillRise: Agentic Reinforcement Learning for Cross-Task Skill Evolution}

\author{
\textbf{Zhiyuan Yao}$^{1}$ \quad
\textbf{Yuxin Chen}$^{2}$ \quad
\textbf{Zhengxi Lu}$^{1}$ \quad
\textbf{Zishan Xu}$^{3}$ \quad
\textbf{Yueqing Sun}$^{4}$ \quad
\textbf{Yifu Guo}$^{}$ \\
\textbf{Yuquan Lu}$^{4}$ \quad
\textbf{Zhengzhou Cai}$^{4}$ \quad
\textbf{Kangning Zhang}$^{3}$ \quad
\textbf{Zhuowen Han}$^{4}$ \quad
\textbf{Zihan Wang}$^{4}$ \\
\textbf{Ziang Ye}$^{4}$ \quad
\textbf{Qi Gu}$^{4,\dagger}$ \quad
\textbf{Xunliang Cai}$^{4}$ \quad
\textbf{Weiwen Liu}$^{3}$ \quad
\textbf{Yongliang Shen}$^{1,\dagger}$ \\
$^{1}$Zhejiang University \quad
$^{2}$National University of Singapore \quad
$^{3}$Shanghai Jiao Tong University \\
$^{4}$Meituan \\
$^{\dagger}$Corresponding authors.
}

\iclrfinalcopy % Uncomment for camera-ready version, but NOT for submission.

\newcommand{\longcatlogo}{%
  \raisebox{-0.21\height}[0pt][0pt]{%
     \includegraphics[height=1.8em, keepaspectratio]{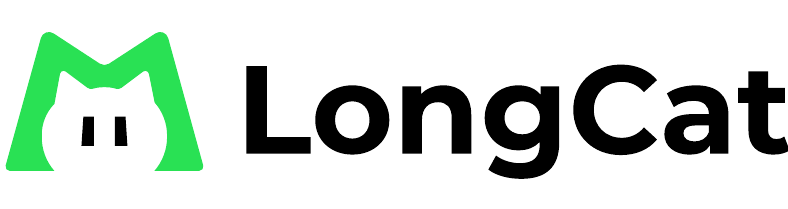}%
  }%
}

\fancypagestyle{firstpage}{%
  \fancyhf{}%
  \fancyhead[L]{\longcatlogo}%
  \renewcommand{\headrulewidth}{0.4pt}%
}

\begin{document}

\maketitle
% \maketitle does not set \thispagestyle, so the first page would otherwise use
% the global "fancy" style; force our first-page header here.
\thispagestyle{firstpage}
% Subsequent pages (page 2 onward): logo on the RIGHT, left empty. (\@maketitle
% sets \lhead to the ICLR notice, so we clear it here.)
\fancyhead{}
\lhead{}
\fancyhead[R]{\longcatlogo}

\begin{abstract}
% The abstract paragraph should be indented 1/2~inch (3~picas) on both left and
% right-hand margins. Use 10~point type, with a vertical spacing of 11~points.
% The word \textsc{Abstract} must be centered, in small caps, and in point size 12. Two
% line spaces precede the abstract. The abstract must be limited to one
% paragraph.
Large language model agents often encounter related yet distinct tasks that
share reusable solution patterns. Yet standard agentic reinforcement learning
treats tasks as independent episodes, while existing approaches to skill
learning either focus on repeated attempts of one task or use pipelines with
multiple stages that entangle extraction, retrieval, and execution. We
introduce \name, a unified reinforcement learning framework for learning skills
across tasks.
\name organizes related instances into progressively challenging sequences and
uses a single policy to alternate between task solving and curating an evolving
skill document passed directly to the next task. Decoupled credit assignment
across tasks supervises solving with the current task outcome and curation with
discounted downstream outcomes. Experiments on ALFWorld, WebShop, and
ScienceWorld show that \name achieves the strongest Pass@1 performance among
the compared methods, with gains over the strongest baseline ranging from 2.3
to 8.5 percentage points. Although trained across distinct tasks, its learned
curation policy remains effective for repeated attempts on the same task.
Further analysis reveals scaling at test time across tasks: performance
improves with longer sequences of related tasks even when each task is
attempted only once. This trend suggests that \name reuses transferable skills
across tasks rather than benefiting from repeated sampling of the same task.
\name further retains strong performance while substantially reducing the
runtime overhead of skill learning pipelines with multiple stages. Together,
these results provide a simple and efficient training paradigm for LLM agents
to extract, refine, and reuse transferable skills across tasks.
Our code is publicly available at \url{https://github.com/Within-yao/SkillRise}.

\end{abstract}

\section{Introduction}

%大语言模型（LLM）智能体正被日益广泛地应用于复杂的长程任务。在实际部署中，智能体面对的往往不是孤立的单个任务，而是一系列相关但不同的任务，它们共享底层规律与可复用的解决模式。然而，标准的智能体强化学习通常将每个任务视为独立的 episode，任务结束后便丢弃交互中获得的经验，导致智能体在后续任务中反复从头探索。理想的智能体不仅应当解决当前任务，还应从交互经验中提炼可迁移的技能，并在后续任务中持续复用和完善这些技能，从而随经验积累而不断提升。

%近期工作开始利用强化学习训练智能体从交互轨迹中提炼可复用的经验或技能。一类方法反复尝试同一任务，并总结先前轨迹以改进后续尝试；但所得知识容易绑定具体实例，也需要额外的交互开销，因而并未直接面向跨任务迁移。另一类方法维护外部经验库或技能库，通过更新、检索和选择技能来指导后续任务。然而，下游表现同时受到技能提取、检索和执行等多个环节影响，使任务结果难以直接判断模型所提取的技能是否真正有效。

%上述局限促使我们以一种简洁、端到端的形式重新建模跨任务技能学习。我们将一系列相关但不同的任务组织为连续序列，使智能体能够从前序任务中提炼技能并用于后续任务。在这一过程中，同一策略根据当前任务的交互轨迹更新技能文档，并将其直接用于下一个任务，从而避免复杂的外部技能管理链路。由于任务之间共享规律但实例各异，后续任务的表现能够自然检验技能的迁移价值，使模型联合学习技能的生成与使用。为了有效优化这两类行为，我们进一步认为，监督信号应与其在任务序列中的作用相对应，其中任务求解由当前任务结果监督，技能更新则由后续任务表现评价。

%基于上述思路，我们提出 XMeta，一个用于跨任务技能学习的端到端强化学习框架。XMeta 首先从同一任务族中选取相似但不同的任务实例，并按照难度将其组织为逐步递进的序列，使前序任务中获得的经验能够为后续任务提供有效基础。沿着该序列，单一策略交替执行任务求解与技能更新，在当前技能文档的指导下完成任务，再根据所得轨迹修订该文档，并将其用于下一个任务实例。为有效优化这两类行为，XMeta 采用解耦的跨任务信用分配，使任务求解仅接收当前任务奖励，而技能更新接收后续任务奖励的折扣回报。此外，XMeta 在相同序列阶段和相同行为角色的多条 trial 之间分别计算组相对优势，避免两类学习信号相互干扰。通过这一训练过程，模型能够在解决任务的同时持续提炼、修订和复用可迁移技能。

%我们在 ALFWorld、WebShop 和 ScienceWorld 上评估了 \name。在这三个智能体基准上，\name 均持续优于基于提示的方法和标准强化学习基线。相较于在同一任务上反复尝试的方法，\name 的优势表明其提取的技能能够在不同任务实例之间迁移。更重要的是，\name 展现出跨任务测试时扩展能力，即随着测试阶段经历的相关任务序列变长，模型能够持续完善技能文档并获得更好的表现，而其他方法没有表现出这一增长趋势。这些结果表明，\name 能够持续提取、完善并复用可迁移技能，并随着接触更多任务而不断提升能力。

Large language model agents are increasingly deployed to solve complex, long-horizon tasks~\citep{anthropic2025claude45,gpt5,guo2025deepseekr1,xie2024osworld}. 
In practice, they often encounter streams of related yet distinct tasks that share underlying regularities and reusable solution patterns~\citep{zheng2025lifelongagentbench,ouyang2026skillos}. 
However, standard agentic RL typically treats each task as an independent episode, discarding the experience acquired during interaction and forcing the agent to repeatedly explore from scratch~\citep{qiu2026autorefine}. 
An ideal agent should not only solve the task at hand, but also extract transferable skills from experience and continually reuse and refine them across subsequent tasks, becoming increasingly capable over time.

Recent work has begun to use reinforcement learning to train agents to extract reusable experience or skills from interaction trajectories~\citep{zhang2026memskill,shi2026skill1,wang2026sage}. 
One line repeatedly attempts the same task and summarizes previous trajectories to improve subsequent attempts~\citep{jiang2025meta}; 
however, the resulting knowledge can remain instance-specific, requires additional interactions, and is not directly optimized for cross-task transfer. 
Another line maintains an external experience or skill bank, updating it with
newly acquired knowledge and retrieving relevant skills to guide future
tasks~\citep{xia2026skillrl,wu2025evolver}.
However, because downstream performance is jointly determined by skill extraction, retrieval, and execution, it is difficult to attribute success or failure to the quality of the extracted skills themselves.

These limitations motivate a simple, end-to-end formulation of cross-task skill learning. We consider a sequence of related yet distinct tasks, in which an agent extracts skills from earlier interactions and applies them to subsequent tasks. 
Within this sequence, the same policy incorporates each interaction trajectory into an evolving skill document and directly uses the document for the next task, avoiding a complex external skill-management pipeline. 
Since the tasks share common regularities while differing in their specific instances, performance on later tasks provides a natural measure of skill transferability and enables the policy to jointly learn to produce and use transferable skills. 
To optimize these two behaviors effectively, we further argue that supervision should follow their temporal roles, with task solving evaluated by the current task outcome and skill curation by performance on subsequent tasks.

Building on this formulation, we introduce \name, an end-to-end reinforcement learning framework for cross-task skill learning.
\name first constructs progressively challenging task sequences by selecting similar yet distinct instances from the same task family and ordering them by difficulty, allowing experience from earlier tasks to support later ones.
During each rollout, a single policy alternates between solving the current task with an evolving skill document and curating the document based on the resulting trajectory before proceeding to the next task instance.
\name employs decoupled cross-task credit assignment that assigns the current task reward to task solving and a discounted return over subsequent task rewards to skill curation.
Group-relative advantages are then computed over trials sharing the same task group, sequence stage, and behavioral phase.
Together, these designs enable the policy to continually extract, refine, and reuse transferable skills while solving a sequence of tasks.

We evaluate \name on ALFWorld, WebShop, and ScienceWorld. Across all three agentic benchmarks, \name consistently outperforms prompting-based methods and standard reinforcement learning baselines. 
Its advantage over methods that repeatedly attempt the same task indicates that the extracted skills transfer across different task instances. 
More importantly, \name exhibits cross-task test-time scaling, improving as it encounters longer sequences of related tasks at test time and progressively refines its skill document. 
Competing methods do not show this growth trend. 
These results show that \name learns a transferable mechanism for cross-task self-improvement, allowing skills acquired from earlier tasks to systematically improve performance on later ones.

Overall, our contributions can be summarized as follows:
\begin{itemize}
    \item We formulate cross-task skill learning over ordered sequences of
    related yet distinct task instances, where later-task outcomes directly
    evaluate whether skills curated from earlier interactions transfer to
    subsequent tasks.

    \item We develop \name, an end-to-end reinforcement learning framework that
    uses a single policy to solve tasks and curate an evolving skill document,
    together with decoupled cross-task credit assignment and role-aware
    group-relative optimization.

    \item Extensive experiments on ALFWorld, WebShop, and ScienceWorld show
    that \name achieves the best overall results among the compared methods.
    Further analyses demonstrate within-task generalization and cross-task
    test-time scaling, and show that its compact end-to-end design offers a
    better balance of effectiveness and efficiency than multi-stage
    skill-learning pipelines.
\end{itemize}

% 我们在 Minesweeper、Sokoban、ScienceWorld 和 WebShop 四个具有代表性的长程任务基准上评估了 VEL。实验结果表明，VEL 在四个基准上均稳定优于基于提示的方法、标准强化学习基线以及现有经验学习方法。通过强化智能体从先前尝试中提炼有效经验的能力，VEL 平均带来了 40\% 的性能提升，展现出有效的测试时适应能力。此外，仅在单一场景中接受 VEL 训练的模型也能够良好地泛化到未见过的任务和环境中。更令人惊讶的是，当我们将连续学习过程从“在同一任务上进行多次尝试”扩展到“依次尝试相关但不同的任务”时，VEL 仍然能够取得显著的性能提升。这些结果表明，VEL 学到的并非针对特定任务的经验记忆，而是一种可迁移、可扩展的经验学习能力，为构建长程自主智能体提供了具有潜力的基础。

\section{Preliminaries}
\label{sec:preliminaries}

We consider an LLM agent that interacts with an environment to solve a task
instruction $x$ sampled from a task distribution $\mathcal{D}$. At each step
$t$, the agent receives an observation $o_t\in\mathcal{O}$ and selects an action
$a_t\in\mathcal{A}$ according to
\begin{equation}
    a_t \sim \pi_\theta(\cdot \mid x,h_t),
\end{equation}
where $\pi_\theta$ denotes the agent policy and
$h_t=(o_0,a_0,\ldots,a_{t-1},o_t)$ is the interaction history available to the
agent. After executing $a_t$, the environment returns the next observation
$o_{t+1}$.

An episode produces a trajectory
$\tau=(o_0,a_0,\ldots,a_{T-1},o_T)$ and terminates when the task is completed or
the interaction budget is exhausted. The trajectory is evaluated by a
verifiable outcome reward $R(\tau)$. The standard reinforcement learning
objective is to maximize the expected task reward:
\begin{equation}
    \mathcal{J}(\theta)
    =
    \mathbb{E}_{x\sim\mathcal{D},\,
    \tau\sim\pi_\theta(\cdot\mid x)}
    \left[R(\tau)\right].
    \label{eq:rl-objective}
\end{equation}
This objective treats each task independently; in the following section, we
extend it to a sequence of related tasks connected through an evolving skill
state.

\section{Method}

\begin{figure}[t]
    \centering
    \includegraphics[width=\linewidth,pagebox=cropbox]{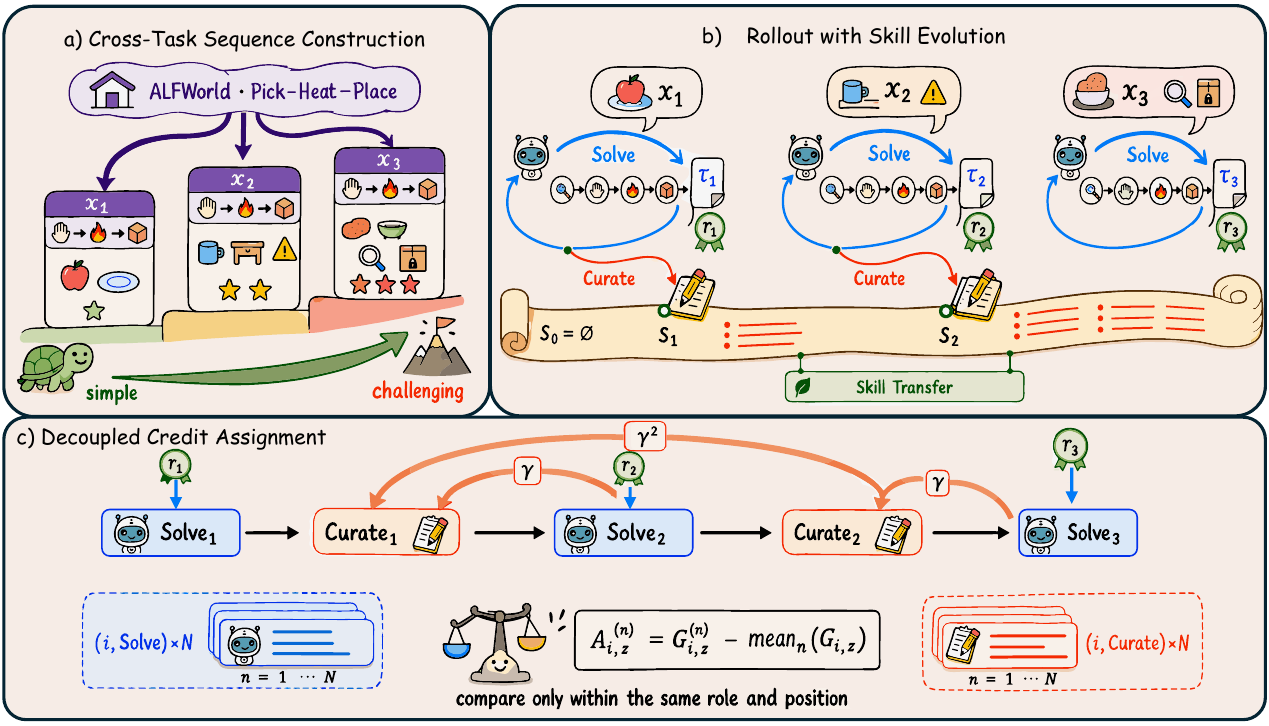}
    \caption{Overview of \name. (a) Related task instances are organized from
    simple to challenging. (b) A shared policy alternates between solving each
    task and curating an evolving skill document, which serves as the sole
    information channel across tasks. (c) Task solving receives the current
    task reward, while skill curation receives discounted rewards from
    subsequent tasks; group-relative advantages are computed among trials with
    the same role and sequence position.}
    \label{fig:method-overview}
\end{figure}

Figure~\ref{fig:method-overview} presents an overview of \name. The framework
first organizes related task instances into a progressive sequence, then uses a
single policy to alternate between task solving and skill curation along the
sequence. The two roles are optimized with temporally aligned learning signals
to jointly improve task-solving and cross-task skill-learning capabilities.

\subsection{Cross-Task Sequence Construction}
\label{sec:cross-task-sequence}

Standard agentic reinforcement learning treats each task as an independent
episode, preventing skills learned from one task from being transferred to
another. To create a learning signal for cross-task skill transfer, we organize
tasks into ordered sequences
\begin{equation}
    \mathbf{x}=(x_1,\ldots,x_K)\sim\mathcal{G},
    \label{eq:cross-task-sequence}
\end{equation}
where $\mathcal{G}$ is a distribution over sequences of $K$ distinct instances
from the same task family. Although these instances differ in their concrete
entities and goals, they share common interaction routines that can be
abstracted into transferable skills. Consequently, skills learned from earlier
tasks can guide later tasks, while performance on the later tasks provides a
direct measure of whether those skills transfer beyond a single instance.

To construct the sequences, we first partition tasks by environment-provided
family metadata, and then greedily group diverse instances within each family.
We arrange the selected tasks from relatively simple to more complex according
to their task attributes. For example, we group WebShop tasks by product
category and order distinct requests by the number of required attributes and
options. The resulting tasks are executed sequentially and connected through
an evolving skill document.

\subsection{Cross-Task Rollout with Skill Evolution}
\label{sec:skill-evolving-rollout}

Given a cross-task sequence $\mathbf{x}$, \name performs a sequential rollout
while maintaining a single textual skill document $S_i$ that summarizes the
transferable skills acquired through the first $i$ tasks. Each trial starts
from an empty document $S_0=\varnothing$. For the $i$-th task, the policy first
uses the current document to generate a task-solving trajectory and receive
its task reward:
\begin{equation}
    \tau_i \sim \pi_\theta(\cdot\mid x_i,S_{i-1}),
    \qquad
    r_i=R(\tau_i).
    \label{eq:skill-conditioned-solving}
\end{equation}
For every task except the last, the same policy then switches from task solving
to skill curation and produces a revised document:
\begin{equation}
    S_i
    \sim
    \pi_\theta(\cdot\mid S_{i-1},\tau_i,r_i),
    \qquad i<K.
    \label{eq:skill-curation}
\end{equation}
The two phases share the same policy parameters and differ only in their
role-specific instructions.

During curation, the policy rewrites the complete document: it preserves useful
skills from $S_{i-1}$, consolidates successful procedures or failure modes
revealed by $\tau_i$, and removes instance-specific details. Only the revised
document $S_i$ is provided to task $x_{i+1}$; earlier task trajectories are not
carried into its solving context. Hence, $S_i$ is the sole information channel
across tasks, and its utility is reflected by performance on subsequent tasks.
During training, we sample $N$ independent trials for each sequence, with every
trial starting from an empty document and evolving its own skill document.

\subsection{Decoupled Credit Assignment and Optimization}
\label{sec:decoupled-optimization}

\paragraph{Decoupled credit assignment.}
Task solving and skill curation play different temporal roles. A solving
trajectory can be evaluated directly by the current task reward, whereas the
revised skill document is produced only after the task terminates and can
influence only subsequent tasks. Assigning the same sequence-level return to
both phases would therefore conflate task-solving quality with skill
transferability. Let $n\in\{1,\ldots,N\}$ index the independent trials and
$z\in\{\mathrm{solve},\mathrm{curate}\}$ denote the phase. We assign the
phase-level return
\begin{equation}
    G_{i,z}^{(n)}
    =
    \begin{cases}
        r_i^{(n)},
        & z=\mathrm{solve},\\[2pt]
        \displaystyle\sum_{j=i+1}^{K}
        \gamma^{\,j-i}r_j^{(n)},
        & z=\mathrm{curate},\ i<K,
    \end{cases}
    \label{eq:decoupled-return}
\end{equation}
where $\gamma\in[0,1]$ is the cross-task discount factor. Thus, solving is
supervised only by the current task outcome, while curation is evaluated by
the discounted outcomes of later tasks. The discount places more credit on
nearby tasks, whose performance is more directly affected by the revised
document.

\paragraph{Role-aware group-relative optimization.}
For a fixed sequence, task position $i$, and phase $z$, we compare the returns
of its $N$ trials:
\begin{equation}
    \overline{G}_{i,z}
    =
    \frac{1}{N}\sum_{n=1}^{N}G_{i,z}^{(n)},
    \qquad
    A_{i,z}^{(n)}
    =
    G_{i,z}^{(n)}-\overline{G}_{i,z}.
    \label{eq:role-aware-advantage}
\end{equation}
Grouping by both task position and phase ensures that solving and curation
returns never serve as baselines for one another, while also accounting for
differences across positions in the sequence.

Let $\rho_{i,z,t}^{(n)}(\theta)$ denote the token-level importance ratio between
the current and old policies for token $t$. We jointly optimize all solving and
curation responses using the clipped objective
\begin{equation}
    \mathcal{J}(\theta)
    =
    \mathbb{E}_{n,i,z,t}
    \left[
        \min\left(
            \rho_{i,z,t}^{(n)}(\theta) A_{i,z}^{(n)},
            \operatorname{clip}\!\left(
                \rho_{i,z,t}^{(n)}(\theta),1-\epsilon,1+\epsilon
            \right) A_{i,z}^{(n)}
        \right)
    \right].
    \label{eq:xmeta-objective}
\end{equation}
where $\epsilon$ is the clipping range. The phase-level advantage is applied to
all policy-generated tokens in the corresponding phase. Since the two phases
share the same policy, this objective jointly improves task solving and skill
curation while keeping their learning signals separated.

\section{Experiments}

\subsection{Experimental Setup}

\paragraph{Benchmarks.}
We evaluate on three interactive text environments. ALFWorld
\citep{shridhar2020alfworld} requires an embodied household agent to navigate
and manipulate objects; we use its TextWorld version and report results for all
six task families: Pick, Look, Clean, Heat, Cool, and Pick2. WebShop
\citep{yao2022webshop} requires an agent to search for, configure, and purchase
a product that satisfies a natural-language request. We use the 1,000-product
version with synthetic goals. ScienceWorld \citep{wang2022scienceworld}
evaluates grounded scientific reasoning through interactive experiments. We evaluate on 128 held-out task instances from
each environment.

\paragraph{Baselines.}
We compare against three prompting baselines: Zero-shot, ReAct
\citep{yao2023react}, and Reflexion \citep{shinn2023reflexion}. We further
include four task-independent RL algorithms: PPO \citep{schulman2017proximal}, RLOO
\citep{ahmadian2024back}, GRPO \citep{shao2024deepseekmath}, and GiGPO
\citep{feng2026group}. Finally, we compare with LaMer
\citep{jiang2025meta}, a Meta-RL baseline that adapts across repeated attempts
of the same task through textual reflection.

\paragraph{Implementation Details.}
All methods use Qwen3-1.7/4B \citep{yang2025qwen3} as the backbone. For \name, each
training batch contains 16 sequences, each with $K=3$ tasks and $N=8$
independent trials. LaMer also uses eight trials with three attempts, whereas
the task-independent RL baselines use 24 rollouts per task. Thus, every
trainable method processes the same 384 task plays per update. We use an actor
learning rate of $1\times10^{-6}$, a mini-batch size of 128, a maximum response
length of 1,024 tokens, and train for up to 150 updates. For \name, the
cross-task discount factor is set to $\gamma=0.6$. Evaluation uses sampling
with temperature 0.7. Training is implemented with FSDP and vLLM, and all
experiments are conducted on eight NVIDIA H800 GPUs.

\begin{table*}[t]
    \centering
    \caption{
    Pass@1 success rate (\%) of Qwen3-4B on ALFWorld, WebShop, and
    ScienceWorld. Best results are shown in bold and second-best results are
    underlined.
    }
    \label{tab:main-results}
    \resizebox{\textwidth}{!}{
    \begin{tabular}{lccccccccc}
        \toprule
        & \multicolumn{7}{c}{ALFWorld}
        & \multicolumn{1}{c}{WebShop}
        & \multicolumn{1}{c}{ScienceWorld} \\
        \cmidrule(lr){2-8}
        \cmidrule(lr){9-9}
        \cmidrule(lr){10-10}
        Method
        & Pick & Look & Clean & Heat & Cool & Pick2 & Avg.
        & SR & SR \\
        \midrule
        Zero-shot
        & 1.5 & 6.1 & 3.6 & 4.2 & 2.8 & 8.6 & 4.4
        & 1.4 & 0.8 \\
        ReAct
        & 45.5 & 30.3 & 21.4 & 12.5 & 18.1 & 23.5 & 25.0
        & 2.1 & 7.8 \\
        Reflexion
        & 74.2 & 39.4 & 19.0 & 25.0 & 13.9 & 13.6 & 28.9
        & 2.7 & 7.8 \\
        \midrule
        PPO
        & 82.1 & 70.0 & 57.9 & 52.9 & 50.0 & \textbf{72.0} & 67.2
        & 74.2 & 39.1 \\
        RLOO
        & 88.2 & 37.5 & 96.1 & 61.9 & \underline{80.0} & 42.1 & 74.2
        & 76.6 & 43.0 \\
        GRPO
        & \underline{94.1} & 75.0 & \underline{96.2} & 66.7
        & 75.0 & 52.6 & 79.7
        & 75.8 & 42.2 \\
        GiGPO
        & 90.3 & \textbf{88.2} & \textbf{96.4} & \textbf{83.3}
        & 75.0 & 60.0 & \underline{83.6}
        & \underline{77.3} & \underline{46.1} \\
        LaMer
        & 88.2 & 62.5 & \underline{96.2} & 71.4
        & 75.0 & 42.1 & 76.7
        & 74.2 & 41.4 \\
        \midrule
        \name
        & \textbf{100.0} & \underline{87.5} & 88.5 & \underline{81.0}
        & \textbf{80.8} & \underline{68.4} & \textbf{85.9}
        & \textbf{84.4} & \textbf{54.6} \\
        \bottomrule
    \end{tabular}
    }
\end{table*}

\begin{table*}[t]
    \centering
    \caption{
    Pass@1/2/3 success rates (\%) of Qwen3-4B on ALFWorld, WebShop, and
    ScienceWorld. Best results are shown in bold and second-best results are
    underlined.
    }
    \label{tab:passk-results}
    \resizebox{\textwidth}{!}{
    \begin{tabular}{lccccccccc}
        \toprule
        & \multicolumn{3}{c}{ALFWorld}
        & \multicolumn{3}{c}{WebShop}
        & \multicolumn{3}{c}{ScienceWorld} \\
        \cmidrule(lr){2-4}
        \cmidrule(lr){5-7}
        \cmidrule(lr){8-10}
        Method
        & Pass@1 & Pass@2 & Pass@3
        & Pass@1 & Pass@2 & Pass@3
        & Pass@1 & Pass@2 & Pass@3 \\
        \midrule
        Zero-shot
        & 4.4 & 9.4 & 12.5
        & 1.4 & 2.1 & 2.3
        & 0.8 & 1.6 & 2.3 \\
        ReAct
        & 25.0 & 45.3 & 57.0
        & 2.1 & 4.5 & 4.5
        & 7.8 & 10.2 & 12.5 \\
        Reflexion
        & 28.9 & 36.7 & 42.2
        & 2.7 & 3.3 & 3.5
        & 7.8 & 12.5 & 17.2 \\
        \midrule
        PPO
        & 67.2 & 69.5 & 73.4
        & 74.2 & 75.8 & 75.8
        & 39.1 & 45.3 & 47.7 \\
        RLOO
        & 74.2 & 80.5 & 82.8
        & 76.6 & 78.1 & 78.9
        & 43.0 & 48.4 & 52.3 \\
        GRPO
        & 79.7 & 81.3 & 82.8
        & 75.8 & 76.6 & 78.1
        & 42.2 & 50.0 & 50.0 \\
        GiGPO
        & \underline{83.6} & \underline{86.7} & 89.1
        & \underline{77.3} & 79.7 & 80.5
        & \underline{46.1} & \underline{50.8} & \underline{55.5} \\
        LaMer
        & 76.7 & \underline{86.7} & \underline{91.4}
        & 74.2 & \underline{87.5} & \underline{94.5}
        & 41.4 & 46.1 & 46.8 \\
        \midrule
        \name
        & \textbf{85.9} & \textbf{91.4} & \textbf{92.2}
        & \textbf{84.4} & \textbf{93.8} & \textbf{96.1}
        & \textbf{54.6} & \textbf{58.5} & \textbf{61.0} \\
        \bottomrule
    \end{tabular}
    }
\end{table*}

\subsection{Main Results}

\paragraph{Overall Performance.}
As shown in Table~\ref{tab:main-results}, \name achieves the strongest overall
Pass@1 performance across all three benchmarks. It obtains 85.9\% on ALFWorld,
84.4\% on WebShop, and 54.6\% on ScienceWorld, outperforming the strongest
baseline, GiGPO, by 2.3, 7.1, and 8.5 percentage points, respectively. On
ALFWorld, \name ranks first or second in five of the six task families, leading
to the best overall success rate despite substantial variation across task
types. We attribute these gains to cross-task training, where related but
distinct instances repeatedly expose the policy to different realizations of
the same underlying interaction routines. This structured training signal makes
transferable regularities easier to identify and reinforce across tasks,
accelerating their internalization into the model parameters and enabling the
policy to acquire reusable task-solving skills more efficiently.

\paragraph{Generalization to Within-Task Adaptation.}
Although \name learns to curate skills from sequences of distinct tasks during
training, we evaluate Pass@2 and Pass@3 by retrying the same held-out task while
carrying forward the evolving skill document. This protocol tests whether the
skill-learning strategy acquired across tasks can generalize to within-task
adaptation. As shown in Table~\ref{tab:passk-results}, \name achieves the best
Pass@2 and Pass@3 on all three benchmarks. At Pass@3, it reaches 92.2\%,
96.1\%, and 61.0\% on ALFWorld, WebShop, and ScienceWorld, exceeding the
strongest baseline by 0.8, 1.6, and 5.5 percentage points, respectively.
Notably, it also outperforms LaMer, which is trained explicitly on repeated
attempts of the same task, by 0.8, 1.6, and 14.2 points. These results suggest
that the learned curation policy captures a general strategy for converting
interaction outcomes into actionable skills, which transfers beyond its
cross-task training setting to guide subsequent attempts on the same task.

\begin{wrapfigure}{r}{0.5\textwidth}
    \vspace{-15pt}
    \centering
    \includegraphics[width=0.98\linewidth]{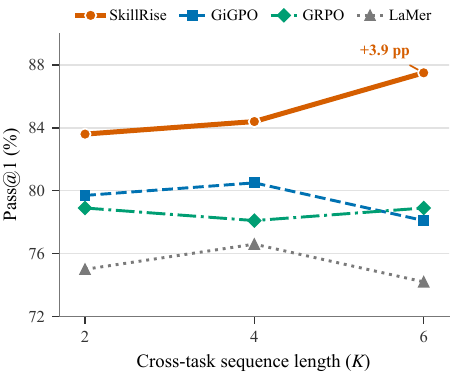}
    \caption{Cross-task test-time scaling on ALFWorld. The same 128 held-out
    tasks are partitioned into related sequences of length $K$, with each task
    attempted once.}
    \label{fig:test-time-scaling}
    \vspace{-10pt} % 微调底部间距，节省空间
\end{wrapfigure}

\section{Analysis}
\label{sec:analysis}

\subsection{Cross-Task Test-Time Scaling}
\label{sec:test-time-scaling}

We investigate whether the cross-task skill-learning capability acquired during
training scales with the amount of related task context available at test time.
We partition the same held-out task set into sequences of
$K\in\{2,4,6\}$ tasks and allow \name to carry forward and update its skill
document within each sequence. As shown in
Figure~\ref{fig:test-time-scaling}, \name improves monotonically over the
evaluated sequence lengths, from 83.6\% at $K=2$ to 87.5\% at $K=6$. In
contrast, LaMer, GiGPO, and GRPO show no sustained improvement as $K$ grows,
and the margin over the strongest baseline widens from 3.9 to 8.6 percentage
points. Since every task receives only one attempt, this gain does not result
from additional sampling on the same instance. Instead, the trend suggests
that the trained policy can transform interactions with earlier tasks into
transferable skills for later tasks, with the benefit becoming more pronounced
as the skill document is progressively refined over longer sequences.

\subsection{Performance across Model Sizes}
\label{sec:model-size}

\begin{table}[t]
    \centering
    \caption{Pass@1 success rate (\%) on ALFWorld across Qwen3 model sizes.
    $\Delta$ denotes the improvement from 1.7B to 4B. Best results are shown
    in bold.}
    \label{tab:model-size}
    \setlength{\tabcolsep}{13pt}
    \begin{tabular}{lccc}
        \toprule
        & \multicolumn{2}{c}{Qwen3} & \\
        \cmidrule(lr){2-3}
        Method & 1.7B & 4B & $\Delta$ \\
        \midrule
        GRPO  & 75.0 & 79.7 & $+4.7$ \\
        LaMer & 73.4 & 76.7 & $+3.3$ \\
        \midrule
        \name & \textbf{78.1} & \textbf{85.9} & $\mathbf{+7.8}$ \\
        \bottomrule
    \end{tabular}
\end{table}

We further evaluate whether the gains of \name extend beyond a single model
size. As shown in Table~\ref{tab:model-size}, \name achieves the strongest
Pass@1 performance at both evaluated scales. With Qwen3-1.7B, \name reaches
78.1\%, exceeding the strongest baseline by 3.1 percentage points. With
Qwen3-4B, it further improves to 85.9\%, widening this margin to 6.2 points.
Thus, the advantage of cross-task skill learning is already present with the
smaller backbone and becomes more pronounced at the larger scale.

All three methods benefit from increasing the backbone size, but the magnitude
of improvement differs substantially. \name gains 7.8 points from 1.7B to 4B,
compared with 4.7 points for GRPO and 3.3 points for LaMer. Unlike
task-independent optimization, \name must jointly learn to solve the current
task, identify regularities shared across related tasks, and curate them into a
reusable skill document. The larger gain suggests that additional model
capacity is particularly useful for this combination of task execution and
cross-task abstraction. Overall, these results show that the effectiveness of
\name is not confined to a single parameter scale and that, within the
evaluated range, its advantage increases as the backbone grows.

\subsection{Ablation Studies}
\label{sec:ablation}

\begin{figure}[t]
    \centering
    \includegraphics[width=0.95\linewidth]{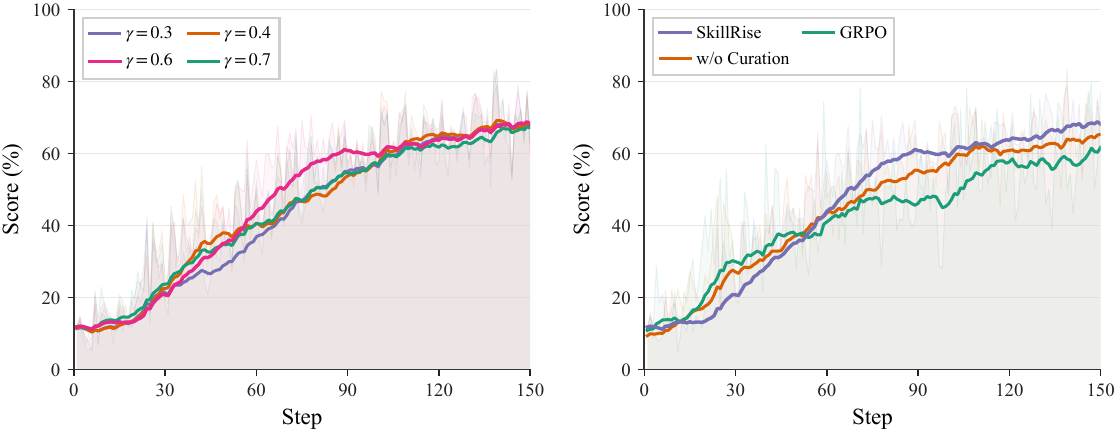}
    \caption{Training dynamics on ALFWorld in terms of training reward
    (\%). Thin lines show the raw rewards, while thick lines show the smoothed
    rewards. Left: sensitivity to the cross-task discount factor. Right:
    comparison of \name, its no-curation variant, and GRPO.}
    \label{fig:ablation-training}
\end{figure}

\paragraph{Sensitivity to Cross-Task Discounting.}
We vary the cross-task discount factor over
$\gamma\in\{0.3,0.4,0.6,0.7\}$ while keeping the remaining settings fixed.
As shown in the left panel of Figure~\ref{fig:ablation-training}, all four
variants follow closely aligned learning trajectories, exhibiting comparable
improvements throughout training and converging within approximately one
percentage point of one another. This consistency shows that the gains of
\name do not rely on carefully tuning the relative weight assigned to future
tasks. Instead, the cross-task learning signal remains effective over a broad
range of discount factors, indicating that the proposed credit assignment is
robust to reasonable variations in how strongly downstream outcomes supervise
skill curation.

\paragraph{Effect of Skill Curation.}
The no-curation variant retains the same $K=3$ task sequences and environment
interaction budget, but removes the inter-task curation phase and prevents
skill transfer between tasks. As shown in the right panel of
Figure~\ref{fig:ablation-training}, the three methods improve similarly during
the early stage, but \name begins to separate after sufficient cross-task
interactions have accumulated and maintains the strongest training score
thereafter. At the end of training, it leads the no-curation variant by roughly
three percentage points and GRPO by more than six points. The advantage over
the no-curation variant shows that training on related task sequences alone is
insufficient. Explicitly transforming completed trajectories into an evolving
skill document allows useful procedures discovered on earlier tasks to be
consolidated and reused on later ones. The additional improvement over
task-independent GRPO further suggests that cross-task skill accumulation
provides a complementary learning signal beyond optimizing each task in
isolation.

Overall, these ablations confirm the importance of skill curation while
showing that \name remains robust to the choice of cross-task discount factor.
Together, they support the effectiveness of the complete framework.

\subsection{Pipeline Complexity and Training Efficiency}
\label{sec:simplified-pipeline}

\begin{figure}[t]
    \centering
    \includegraphics[width=0.95\linewidth]{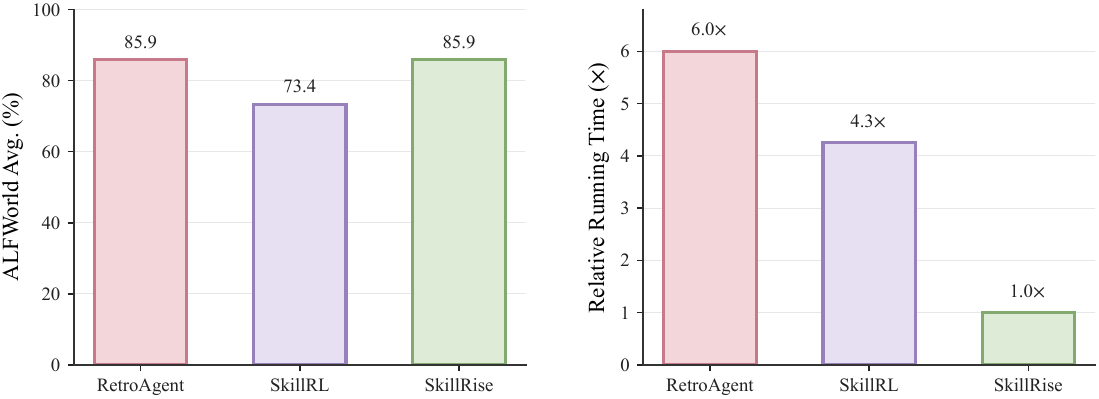}
    \caption{Comparison of skill-learning pipelines on ALFWorld. Left: average
    success rate. Right: relative running time normalized by \name.}
    \label{fig:efficiency-comparison}
\end{figure}

Existing skill-learning methods often decompose skill acquisition into a
multi-stage management pipeline. RetroAgent generates reflections, stores them
in a growing memory buffer, retrieves relevant lessons, and updates their
utilities \citep{zhang2026retroagent}. SkillRL further relies on teacher-based
trajectory distillation, hierarchical SkillBank construction, cold-start SFT,
skill retrieval, and iterative bank refinement \citep{xia2026skillrl}. For our
SkillRL reproduction, we use Gemini-2.5-Pro as the teacher model. In such
pipelines, downstream performance is jointly determined by skill extraction,
storage, retrieval, updating, and execution. This coupling makes it difficult
to attribute success or failure to the learned skill itself, while the
additional stages also introduce substantial computational overhead.

\name instead adopts a compact end-to-end training paradigm: the same policy
solves each task, curates a single sequence-local skill document, and directly
uses it on the next task. This removes the need for an external teacher,
separate memory maintenance, and an additional retrieval module. As shown in
Figure~\ref{fig:efficiency-comparison}, \name achieves an 85.9\% average
success rate on ALFWorld, matching RetroAgent and outperforming SkillRL by
12.5 percentage points. Under the same GPU configuration, RetroAgent and
SkillRL require $6.0\times$ and $4.3\times$ the end-to-end running time of
\name, respectively. Thus, transferable skill learning does not require a long
external management pipeline. Directly supervising skill curation through
subsequent task outcomes provides a clearer learning signal while retaining
strong performance with substantially lower training overhead.

\section{Related Work}

\paragraph{Reinforcement Learning for LLM Agents.}
Reinforcement learning has become a central approach for improving LLMs,
evolving from preference alignment with human feedback to reinforcement
learning with verifiable rewards, which provides scalable supervision for
reasoning and improves training efficiency
\citep{shao2024deepseekmath,guo2025deepseekr1,
yao2026coba,chen2026learning}. As LLMs are increasingly deployed as interactive agents,
agentic RL extends this paradigm from single-turn generation to multi-turn
decision making, where long horizons and delayed outcomes make exploration and
credit assignment substantially harder \citep{zhang2025landscape}.
Existing methods develop hierarchical, step-level, or
milestone-guided credit assignment
\citep{feng2026group,cm2,he2026hierarchy,
wang2026milestone,lu2026sdar}, improve exploration through autonomous environment interaction
and online curricula, including for web navigation and long-context software
engineering \citep{ye2026look,qi2025webrl,wei2026swe,
wang2025ragen}, or provide scalable
training abstractions for heterogeneous agent workflows
\citep{luo2025agent,xi2025agentgym}. These advances demonstrate
that outcome-based supervision can improve long-horizon interaction, but most
methods still optimize independently sampled task episodes. Consequently, how
an agent can learn from the relationship among episodes and turn earlier
interactions into reusable procedural knowledge remains less explored.

\paragraph{Experience and Skill Learning in LLM Agents.}
Learning from past interactions is a central mechanism for self-evolving
agents, with the core challenge being how to transform interaction histories
into reusable knowledge \citep{gao2025survey}.
Concrete representations retain trajectories, examples, or verbal reflections
for later reuse \citep{shinn2023reflexion,zhao2024expel}. More abstract
representations distill histories into guidelines, workflows, or reasoning
strategies that can generalize across instances
\citep{fu2024autoguide,wang2024agent,ouyang2025reasoningbank}.
Recently, editable skill artifacts have emerged as a modular form of
procedural memory, allowing agents to consolidate, revise, and reuse
high-level solution procedures \citep{ni2026trace2skill,yang2026skillopt}.
Reinforcement learning is increasingly used to train different stages of this
skill lifecycle, including skill utilization, internalization, extraction, and
curation \citep{xia2026skillrl,
lu2026skill0,ouyang2026skillos}. LaMer
formulates adaptation as a cross-episode Meta-RL problem: the agent repeatedly
interacts with the same task and conditions later attempts on environment
feedback and textual reflections \citep{jiang2025meta}. For skill evolution,
Evolving-RL jointly optimizes skill extraction and skill-conditioned solving
by evaluating extracted skills on related downstream tasks
\citep{fan2026evolving}. SkillOS keeps the executor frozen and trains a
separate curator to update a skill repository, using later performance over
related task streams as delayed supervision
\citep{ouyang2026skillos}.
These methods manage skill extraction, storage,
retrieval, and execution in different ways. However, multi-stage pipelines can
make it difficult to attribute downstream performance to individual
components.

\section{Conclusion}

We introduced \name, an end-to-end reinforcement learning framework for
cross-task skill learning. It organizes related yet distinct tasks into ordered
sequences, where a single policy alternates between task solving and curating
an evolving skill document. Decoupled credit assignment uses current-task
outcomes for solving and downstream outcomes for skill curation, aligning
supervision with their temporal roles. Across ALFWorld, WebShop, and
ScienceWorld, \name achieves the strongest overall performance among the
compared methods, generalizes from cross-task training to repeated attempts,
and improves with longer related-task sequences at test time. Moreover, the
pipeline comparison shows that \name maintains strong performance while
requiring only one sixth of the running time. Overall, our work provides a
simple and efficient training paradigm for enabling LLM agents to extract,
refine, and reuse transferable skills across tasks.

\section{Limitation}

Our current formulation assumes access to task-family metadata for constructing
sequences of related instances; automatically discovering such relationships
in open-ended task streams remains an important direction. Moreover, due to
computational constraints, our experiments are limited to models of up to
4B parameters, leaving evaluation at larger model scales for future work.
Finally, our experiments focus on three text-based agent benchmarks and
verifiable outcome rewards. Evaluating the framework on broader multimodal,
real-world, and less readily verifiable tasks would further establish its
generality.

\bibliography{iclr2026_conference}
\bibliographystyle{iclr2026_conference}

\appendix
% \section{Appendix}

\clearpage
\section{Prompts}
\label{app:prompts}

\newtcolorbox{promptbox}[1]{
    enhanced, colback=black!2, colframe=black!55,
    coltitle=white, colbacktitle=black!60,
    title={#1}, fonttitle=\bfseries\small,
    fontupper=\scriptsize\ttfamily,
    boxrule=0.7pt, arc=1pt, outer arc=1pt,
    left=6pt, right=6pt, top=4pt, bottom=4pt,
    before skip=3pt, after skip=3pt,
    before upper={\setlength{\parindent}{0pt}\setlength{\parskip}{2pt}}
}
\newcommand{\promptvar}[1]{\texttt{\{#1\}}}
\newcommand{\prompttext}[1]{\texttt{#1}}
\newcommand{\prompttag}[1]{\texttt{\ensuremath{<}#1\ensuremath{>}}}

We reproduce the user-message templates used for task solving and skill
curation. Braced fields are filled at runtime. At the start of each task
sequence, an empty skill document is represented by
\texttt{(The skill document is currently empty. No skills have been distilled
yet.)}

\begin{figure}[!htbp]
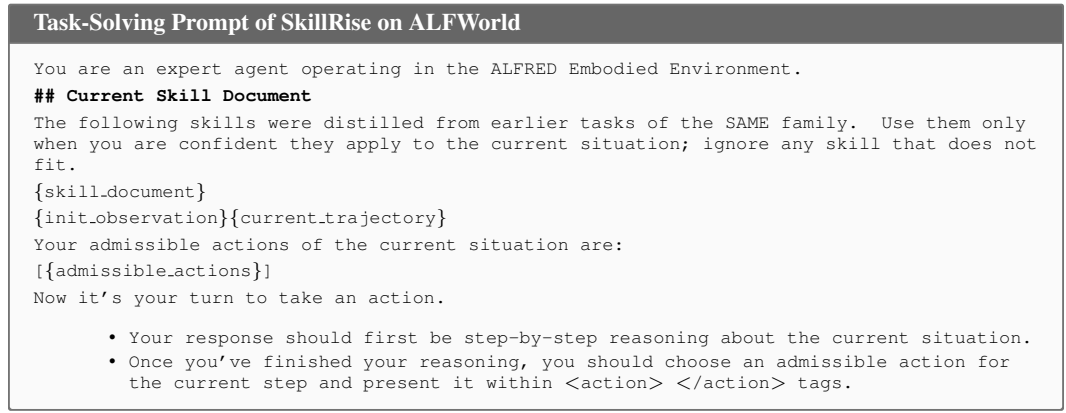

\centering
\begin{promptbox}{Task-Solving Prompt of \name on ALFWorld}
You are an expert agent operating in the ALFRED Embodied Environment.

\textbf{\#\# Current Skill Document}

The following skills were distilled from earlier tasks of the SAME family. Use
them only when you are confident they apply to the current situation; ignore
any skill that does not fit.

\promptvar{skill\_document}

\promptvar{init\_observation}\promptvar{current\_trajectory}

Your admissible actions of the current situation are:

[\promptvar{admissible\_actions}]

Now it's your turn to take an action.

\begin{itemize}
    \setlength{\itemsep}{1pt}
    \setlength{\parskip}{0pt}
    \item Your response should first be step-by-step reasoning about the
    current situation.
    \item Once you've finished your reasoning, you should choose an admissible
    action for the current step and present it within
    \prompttag{action} \prompttag{/action} tags.
\end{itemize}
\end{promptbox}
\caption{\name task-solving prompt for ALFWorld.}
\label{fig:prompt-solve-alfworld}
\end{figure}

\begin{figure}[!htbp]
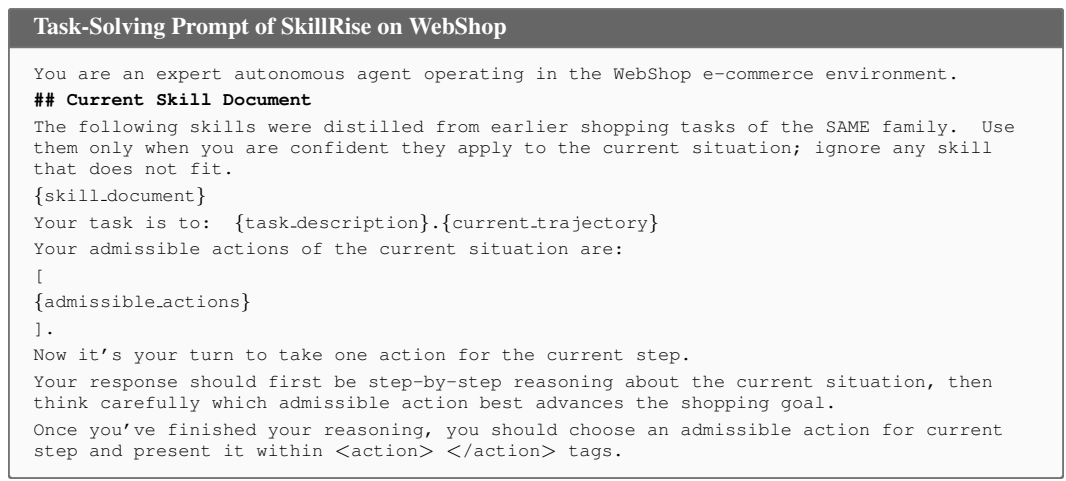

\centering
\begin{promptbox}{Task-Solving Prompt of \name on WebShop}
You are an expert autonomous agent operating in the WebShop e-commerce
environment.

\textbf{\#\# Current Skill Document}

The following skills were distilled from earlier shopping tasks of the SAME
family. Use them only when you are confident they apply to the current
situation; ignore any skill that does not fit.

\promptvar{skill\_document}

Your task is to: \promptvar{task\_description}.\promptvar{current\_trajectory}

Your admissible actions of the current situation are:

[

\promptvar{admissible\_actions}

].

Now it's your turn to take one action for the current step.

Your response should first be step-by-step reasoning about the current
situation, then think carefully which admissible action best advances the
shopping goal.

Once you've finished your reasoning, you should choose an admissible action for
current step and present it within \prompttag{action} \prompttag{/action} tags.
\end{promptbox}
\caption{\name task-solving prompt for WebShop.}
\label{fig:prompt-solve-webshop}
\end{figure}

\begin{figure}[!htbp]
\centering
\begin{promptbox}{Task-Solving Prompt of \name on ScienceWorld}
You are an expert autonomous agent operating in the ScienceWorld environment,
which is a text-based virtual environment centered around accomplishing tasks
from the elementary science curriculum.

\textbf{\#\# Current Skill Document}

The following skills were distilled from earlier tasks of the SAME family. Use
them only when you are confident they apply to the current situation; ignore
any skill that does not fit.

\promptvar{skill\_document}

Your task is to: \promptvar{task\_description}.

Your current observation is:
\promptvar{current\_observation}.\promptvar{current\_trajectory}

Your admissible actions of the current situation are:

[

\promptvar{available\_actions}

].

Now it's your turn to take one action for the current step.

You should first reason step-by-step about the current situation, then think
carefully which admissible action best advances the task goal.

Once you've finished your reasoning, you should choose an admissible action for
current step and present it within \prompttag{action} \prompttag{/action} tags.
\end{promptbox}
\caption{\name task-solving prompt for ScienceWorld.}
\label{fig:prompt-solve-sciworld}
\end{figure}

\begin{figure}[!htbp]
\centering
\begin{promptbox}{Skill-Curation Prompt of \name on ALFWorld}
You are maintaining a SKILL DOCUMENT for an agent solving a family of related
ALFRED household tasks. You have just observed one task attempt. Your job is to
revise the skill document so that it helps the agent solve the LATER tasks in
this same family more reliably.

\textbf{\#\# Old Skill Document}

\promptvar{skill\_document}

\textbf{\#\# Most Recent Task Attempt (observations and the actions the agent
took)}

\promptvar{current\_trajectory}

Outcome of this attempt: \promptvar{outcome}

\textbf{\#\# Your job}

Rewrite the skill document. Start from the old document above and refine it
with what this attempt revealed. Follow these rules strictly:

\begin{itemize}
    \setlength{\itemsep}{1pt}
    \setlength{\parskip}{0pt}
    \item ABSTRACT, do not memorize. Remove task-specific instances: replace
    concrete object names, receptacle ids, and location numbers (e.g.,
    ``soapbottle 1'', ``drawer 3'') with general concepts (``the target
    object'', ``a receptacle'').
    \item Ground every statement in what actually happened in the attempt
    above. Do not invent steps you did not observe.
    \item If the attempt failed, extract the failure point as a concrete
    pitfall.
    \item Be CONCISE. Do not pad with generic advice or vague tips. A short,
    precise document beats a long one. If this attempt revealed little new,
    keep the old document almost unchanged.
    \item Do NOT copy the trajectory verbatim. Distill reusable procedure, not
    history.
\end{itemize}

Suggested (not mandatory) structure---use the headers that fit:

\textbf{\#\# When to use}

\prompttag{high-level situations this skill family applies to}

\textbf{\#\# Workflow}

\prompttag{ordered high-level steps that generalize across instances}

\textbf{\#\# Pitfalls}

\prompttag{failure modes and when to deviate from the default workflow}

Write your reasoning first, then output the FULL revised skill document inside
\prompttag{skill} \prompttag{/skill} tags.
\end{promptbox}
\caption{\name skill-curation prompt for ALFWorld.}
\label{fig:prompt-curate-alfworld}
\end{figure}

\begin{figure}[!htbp]
\centering
\begin{promptbox}{Skill-Curation Prompt of \name on WebShop}
You are maintaining a SKILL DOCUMENT for an agent solving a family of related
WebShop online-shopping tasks. You have just observed one task attempt. Your
job is to revise the skill document so that it helps the agent solve the LATER
tasks in this same family (same product category) more reliably.

\textbf{\#\# Old Skill Document}

\promptvar{skill\_document}

\textbf{\#\# Most Recent Task Attempt (the shopping goal, observations, and the
actions the agent took)}

\promptvar{current\_trajectory}

Outcome of this attempt: \promptvar{outcome}

\textbf{\#\# Your job}

Rewrite the skill document. Start from the old document above and refine it
with what this attempt revealed. Follow these rules strictly:

\begin{itemize}
    \setlength{\itemsep}{1pt}
    \setlength{\parskip}{0pt}
    \item ABSTRACT, do not memorize. Remove task-specific instances: replace
    concrete product names, brands, exact attribute values, and price
    thresholds (e.g., ``rose red'', ``1.6 ounce'', ``lower than 60.00
    dollars'') with general concepts (``the required color/size option'', ``the
    price ceiling'').
    \item Ground every statement in what actually happened in the attempt
    above. Do not invent steps you did not observe.
    \item If the attempt failed, extract the failure point as a concrete
    pitfall.
    \item Be CONCISE. Do not pad with generic advice or vague tips. A short,
    precise document beats a long one. If this attempt revealed little new,
    keep the old document almost unchanged.
    \item Do NOT copy the trajectory verbatim. Distill reusable procedure, not
    history.
\end{itemize}

When useful, the document should capture WebShop-specific procedure:

\begin{itemize}
    \setlength{\itemsep}{1pt}
    \setlength{\parskip}{0pt}
    \item how to formulate a search query that encodes the goal's constraints,
    \item which search result to click (matching category/attributes, not just
    the first),
    \item the order to configure product options (color, size, count, etc.)
    before buying,
    \item verifying the product meets the required attributes AND the price
    ceiling BEFORE \prompttext{click[buy now]},
    \item navigation patterns (when to go back, view
    options/description/features, paginate).
\end{itemize}

Suggested (not mandatory) structure -- use the headers that fit:

\textbf{\#\# When to use}

\prompttag{high-level situations this skill family applies to}

\textbf{\#\# Workflow}

\prompttag{ordered high-level steps that generalize across product instances}

\textbf{\#\# Pitfalls}

\prompttag{failure modes and when to deviate from the default workflow}

Write your reasoning first, then output the FULL revised skill document inside
\prompttag{skill} \prompttag{/skill} tags.
\end{promptbox}
\caption{\name skill-curation prompt for WebShop.}
\label{fig:prompt-curate-webshop}
\end{figure}

\begin{figure}[!htbp]
\centering
\begin{promptbox}{Skill-Curation Prompt of \name on ScienceWorld}
You are maintaining a SKILL DOCUMENT for an agent solving a family of related
ScienceWorld tasks. You have just observed one task attempt. Your job is to
revise the skill document so that it helps the agent solve the LATER tasks in
this same family more reliably.

\textbf{\#\# Old Skill Document}

\promptvar{skill\_document}

\textbf{\#\# Most Recent Task Attempt (observations and the actions the agent
took)}

\promptvar{current\_trajectory}

Outcome of this attempt: \promptvar{outcome}

\textbf{\#\# Your job}

Rewrite the skill document. Start from the old document above and refine it
with what this attempt revealed. Follow these rules strictly:

\begin{itemize}
    \setlength{\itemsep}{1pt}
    \setlength{\parskip}{0pt}
    \item ABSTRACT, do not memorize. Replace concrete instances (specific
    substance names, container colors, room names, numeric thresholds) with
    general concepts (``the target substance'', ``the designated container'').
    \item Ground every statement in what actually happened in the attempt
    above. Do not invent steps you did not observe.
    \item If the attempt failed, extract the failure point as a concrete
    pitfall.
    \item Be CONCISE. A short, precise document beats a long one. If this
    attempt revealed little new, keep the old document almost unchanged.
    \item Do NOT copy the trajectory verbatim. Distill reusable procedure, not
    history.
\end{itemize}

Suggested (not mandatory) structure---use the headers that fit:

\textbf{\#\# When to use}

\prompttag{high-level situations this skill family applies to}

\textbf{\#\# Workflow}

\prompttag{ordered high-level steps that generalize across instances}

\textbf{\#\# Pitfalls}

\prompttag{failure modes and when to deviate from the default workflow}

Write your reasoning first, then output the FULL revised skill document inside
\prompttag{skill} \prompttag{/skill} tags.
\end{promptbox}
\caption{\name skill-curation prompt for ScienceWorld.}
\label{fig:prompt-curate-sciworld}
\end{figure}

\clearpage
\section{Case Study}
\label{app:case}

We show selected actions and abridged curation outputs from the first two tasks
of two ALFWorld training rollouts with sequence length $K=3$. Displayed tasks
are numbered from 1. Each rollout begins with an empty skill document; between
tasks, only the updated document is carried forward, not earlier-task
trajectories. Ellipses mark omitted actions; arrows and comments after
\texttt{\#} are author annotations. Case~1 (Figure~\ref{fig:case-fail}) pairs a
failure-derived rule with the next task conditioned on that rule. Case~2
(Figure~\ref{fig:case-evolve}) shows an incremental document revision. These
examples illustrate the update process rather than isolate a causal effect.

\newtcolorbox{solvebox}[1]{
    enhanced, colback=black!3, colframe=black!55,
    coltitle=white, colbacktitle=black!60,
    title={#1}, fonttitle=\bfseries\small, fontupper=\footnotesize\ttfamily,
    boxrule=0.7pt, arc=1pt, outer arc=1pt,
    left=7pt, right=7pt, top=5pt, bottom=5pt,
    before skip=3pt, after skip=3pt,
}
\newtcolorbox{skillbox}[1]{
    enhanced, colback=blue!4, colframe=blue!55,
    coltitle=white, colbacktitle=blue!55,
    title={#1}, fonttitle=\bfseries\small, fontupper=\footnotesize,
    boxrule=0.7pt, arc=1pt, outer arc=1pt,
    left=7pt, right=7pt, top=5pt, bottom=5pt,
    before skip=3pt, after skip=3pt,
    before upper={\setlength{\parindent}{0pt}\setlength{\parskip}{2pt}}
}
% A centered downward arrow with a short annotation, connecting stacked panels.
\newcommand{\flowarrow}[1]{%
    \begin{center}
    \vspace{-1pt}
    {\Large$\big\downarrow$}\;\raisebox{2pt}{\footnotesize\itshape #1}
    \vspace{-1pt}
    \end{center}%
}

\paragraph{Case 1: distilling a lesson from failure.}
The first task asks for two target instances, but the agent places only one and
fails. The subsequent curation output adds a general instruction to repeat the
procedure for each instance. The next task receives this document, places both
instances of a different object, and succeeds.

\begin{itemize}
    \setlength{\itemsep}{1pt}\setlength{\parskip}{0pt}
    \item \textbf{Task 1}: put \textbf{two} spraybottle in cabinet~~($\times$~failed)
    \item \textbf{Task 2}: put \textbf{two} peppershaker in cabinet~~($\checkmark$~succeeded)
\end{itemize}

\begin{figure}[t]
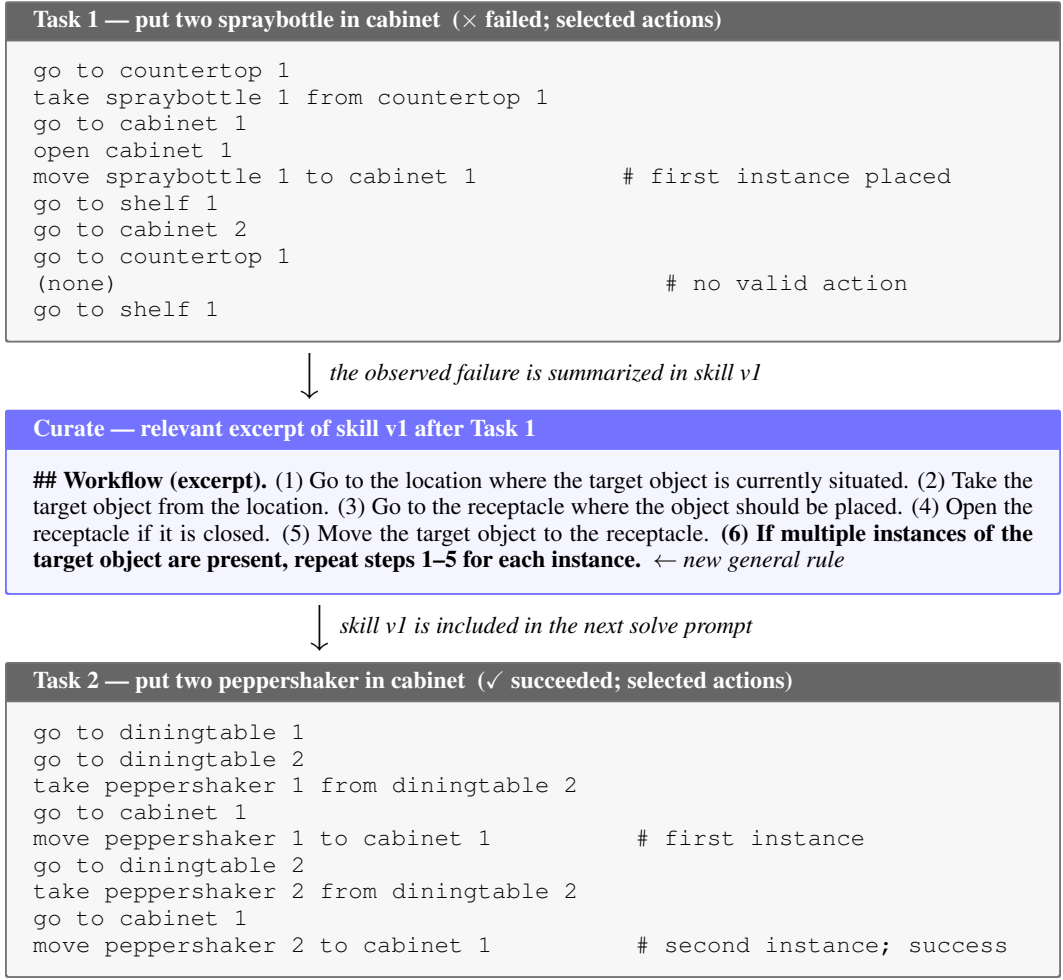

\centering

\begin{solvebox}{Task 1 --- put \textbf{two} spraybottle in cabinet~~($\times$ failed; selected actions)}
go to countertop 1\\
take spraybottle 1 from countertop 1\\
go to cabinet 1\\
open cabinet 1\\
move spraybottle 1 to cabinet 1~~~~~~~~~~\# first instance placed\\
go to shelf 1\\
go to cabinet 2\\
go to countertop 1\\
(none)~~~~~~~~~~~~~~~~~~~~~~~~~~~~~~~~~~~~~~\# no valid action\\
go to shelf 1
\end{solvebox}

\flowarrow{the observed failure is summarized in skill v1}

\begin{skillbox}{Curate --- relevant excerpt of skill v1 after Task 1}
\textbf{\#\# Workflow (excerpt).}
(1) Go to the location where the target object is currently situated.
(2) Take the target object from the location.
(3) Go to the receptacle where the object should be placed.
(4) Open the receptacle if it is closed.
(5) Move the target object to the receptacle.
\textbf{(6) If multiple instances of the target object are present, repeat
steps 1--5 for each instance.}~~$\leftarrow$ \emph{new general rule}
\end{skillbox}

\flowarrow{skill v1 is included in the next solve prompt}

\begin{solvebox}{Task 2 --- put \textbf{two} peppershaker in cabinet~~($\checkmark$ succeeded; selected actions)}
go to diningtable 1\\
go to diningtable 2\\
take peppershaker 1 from diningtable 2\\
go to cabinet 1\\
move peppershaker 1 to cabinet 1~~~~~~~~~~\# first instance\\
go to diningtable 2\\
take peppershaker 2 from diningtable 2\\
go to cabinet 1\\
move peppershaker 2 to cabinet 1~~~~~~~~~~\# second instance; success
\end{solvebox}

\caption{\textbf{Case 1: a rule recorded after failure.} After a failure in
which only one of two required objects is placed, the curation output records a
repeat-for-each-instance rule. The following task is conditioned on this
document and successfully places both instances of a different object.}
\label{fig:case-fail}
\end{figure}

\clearpage
\paragraph{Case 2: incremental skill evolution.}
Both displayed tasks succeed. Task~1 places an object on a destination that
requires no \texttt{open} action, and skill v1 contains a basic placement
workflow. Task~2 requires opening a closed microwave; afterward, the curation
output adds an ``open the destination if closed'' condition to skill v2. This
case documents the revision itself, not downstream use of the new condition.

\begin{figure}[t]
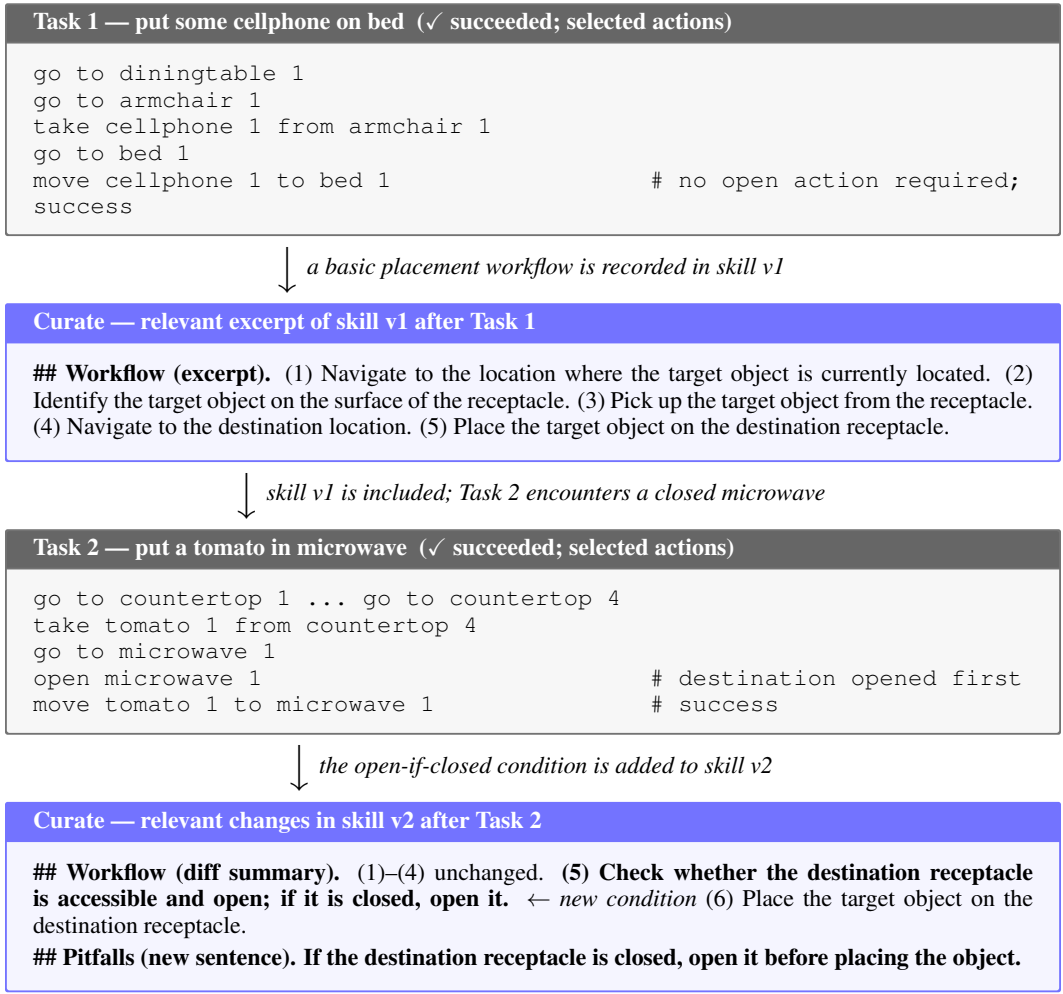

\centering

\begin{solvebox}{Task 1 --- put some cellphone on bed~~($\checkmark$ succeeded; selected actions)}
go to diningtable 1\\
go to armchair 1\\
take cellphone 1 from armchair 1\\
go to bed 1\\
move cellphone 1 to bed 1~~~~~~~~~~~~~~~~~~\# no open action required; success
\end{solvebox}

\flowarrow{a basic placement workflow is recorded in skill v1}

\begin{skillbox}{Curate --- relevant excerpt of skill v1 after Task 1}
\textbf{\#\# Workflow (excerpt).}
(1) Navigate to the location where the target object is currently located.
(2) Identify the target object on the surface of the receptacle.
(3) Pick up the target object from the receptacle.
(4) Navigate to the destination location.
(5) Place the target object on the destination receptacle.
\end{skillbox}

\flowarrow{skill v1 is included; Task 2 encounters a closed microwave}

\begin{solvebox}{Task 2 --- put a tomato in microwave~~($\checkmark$ succeeded; selected actions)}
go to countertop 1 \dots\ go to countertop 4\\
take tomato 1 from countertop 4\\
go to microwave 1\\
open microwave 1~~~~~~~~~~~~~~~~~~~~~~~~~~~\# destination opened first\\
move tomato 1 to microwave 1~~~~~~~~~~~~~~~\# success
\end{solvebox}

\flowarrow{the open-if-closed condition is added to skill v2}

\begin{skillbox}{Curate --- relevant changes in skill v2 after Task 2}
\textbf{\#\# Workflow (diff summary).}
(1)--(4) unchanged.
\textbf{(5) Check whether the destination receptacle is accessible and open;
if it is closed, open it.}~~$\leftarrow$ \emph{new condition}
(6) Place the target object on the destination receptacle.

\textbf{\#\# Pitfalls (new sentence).} \textbf{If the destination receptacle is
closed, open it before placing the object.}
\end{skillbox}

\caption{\textbf{Case 2: an incremental skill revision.} After a successful
task that requires opening the destination, curation adds an ``open if closed''
condition to the existing placement workflow. The added condition uses generic
object and receptacle terms.}
\label{fig:case-evolve}
\end{figure}

\end{document}